\documentclass[pdflatex,sn-mathphys-num]{sn-jnl}

\usepackage{graphicx}%
\usepackage{multirow}%
\usepackage{amsmath,amssymb,amsfonts}%
\usepackage{amsthm}%
\usepackage{mathrsfs}%
\usepackage[title]{appendix}%
\usepackage{xcolor}%
\usepackage{textcomp}%
\usepackage{manyfoot}%
\usepackage{booktabs}%
\usepackage{algorithm}%
\usepackage{algorithmicx}%
\usepackage{algpseudocode}%
\usepackage{listings}%
\usepackage{xcolor, soul}

\usepackage{enumitem}
\usepackage{longtable}
\usepackage{pdflscape}
\usepackage[normalem]{ulem}
\useunder{\uline}{\ul}{}
\usepackage{subcaption}

\begin{document}

\title[Reviewing LLMs]{Decoding Large-Language Models: A Systematic Overview of Socio-Technical Impacts, Constraints, and Emerging Questions.}


\author[1]{\fnm{Zeyneb N.} \sur{Kaya}}\email{zeynebnk@stanford.edu}

\author*[2]{\fnm{Souvick} \sur{Ghosh}}\email{souvick.ghosh@sjsu.edu}

\affil[1]{\orgname{Stanford University}, \orgaddress{\street{450 Jane Stanford Way}, \city{Stanford}, \postcode{94305–2004}, \state{CA}, \country{USA}}}

\affil*[2]{\orgdiv{School of Information}, \orgname{San José State University}, \orgaddress{\street{One Washington Square}, \city{San José}, \postcode{5192-0029}, \state{CA}, \country{USA}}}


\abstract{
There have been rapid advancements in the capabilities of large language models (LLMs) in recent years, greatly revolutionizing the field of natural language processing (NLP) and artificial intelligence (AI) to understand and interact with human language. Therefore, in this work, we conduct a systematic investigation of the literature to identify the prominent themes and directions of LLM developments, impacts, and limitations. Our findings illustrate the aims, methodologies, limitations, and future directions of LLM research. It includes responsible development considerations, algorithmic improvements, ethical challenges, and societal implications of LLM development. Overall, this paper provides a rigorous and comprehensive overview of current research in LLM and identifies potential directions for future development. The article highlights the application areas that could have a positive impact on society along with the ethical considerations. 
}




\maketitle


\section{Introduction}

The capacity of artificial intelligence (AI) models to understand human language holds paramount importance for their interaction with humans and other systems. The emergence of large language models (LLMs) has had a profound impact on the developments of Natural Language Processing (NLP), with models of growing scale. In particular, recent advancements in neural architectures, such as transformer models, have enabled the rapidly expanding capabilities of LLMs, incorporating vast datasets, significant computational power, and an increasing number of parameters. 
With scale, the transition from Language Models (LMs) to LLMs has brought about much greater language capabilities and abilities across tasks. While LMs have largely been limited to basic text processing and specialized applications, LLMs are more generalizable.
LLMs have made notable contributions to language understanding and text generation, handling complex tasks including classification, translation, question-answering, summarization, and information retrieval. Their abilities across a range of diverse tasks and domains, adapting to even low-resource settings~\citep{brown_language_2020}
, demonstrate their transformative generalization capabilities. For instance, models like GPT-3 have showcased remarkable versatility in generating creative content and simulating conversation. The rise of open-source initiatives of LLM architectures and pretrained models has further propelled development. The significant technological impacts of LLMs, enhancing the ability of artificial intelligence systems to communicate with humans, have also had a growing influence on society. Concurrently, these developments raise pressing ethical questions, particularly around data privacy, bias, and the potential for misuse~\citep{bender_dangers_2021}. 
There have been increasing efforts on the release of LLMs, introducing implications of the potential benefits and challenges they pose to our interactions with language and technology.

LLMs have now become the frontier of research and development in NLP and artificial intelligence as a whole. As LLMs grow rapidly with new breakthroughs, applications, and scales, we aim to provide a comprehensive and systematic overview of the directions of LLM research and facilitate further advancements. We summarize the course of existing efforts with LLMs, delving into each component of the model development process and detailing the goals and current capabilities in the field. We further identify the key considerations and gaps to address in LLM development to guide future research. Through these explorations, we aim to answer three guiding research questions (RQs):

\begin{itemize}
    \item RQ1: What are the prominent aims and objectives addressed in LLM research?
    \item RQ2: What are the popular methodologies for advancing the capabilities of LLMs?
    \item RQ3: What are the limitations and ethical considerations of LLM development as identified in contemporary influential research?
\end{itemize}

We investigate these questions through a systematic review characterized by methodological rigor, objectivity, and comprehensive and quantitative statistical analyses of studies, enabling synthesized, evidence-based contributions. 
The rest of the paper is organized as follows: Section 2 provides an overview of the systematic review process, while Section 3 discusses the methodology employed for conducting this review. Section 4 presents several statistics of the reviewed papers. Sections 5, 6, and 7 highlight the purpose, methodology, and limitations of the reviewed papers, respectively. Section 8 answers the research questions, offering insights into future developments, and Section 9 concludes the paper.


\section{Systematic Reviews}
Systematic reviews aim to provide a structured and comprehensive evaluation of existing literature utilizing explicit, replicable methods. Systematic reviews follow a rigorous, meta-analytical approach, employing a standardized procedure for identifying, appraising, and synthesizing relevant studies, culminating in an evidence and statistically-based~\citep{crowther_systematic_2010}
. This approach ensures that the summary of available studies is not only comprehensive but also underpinned by robust methodological rigor.

Owing to their statistical precision, systematic reviews are particularly useful for researchers who rely on evidence-based decision-making. Initially popularized in the health sciences to enhance the reliability and validity of medical research~\citep{manchikanti_evidence-based_2009, tawfik_step_2019}, the practice of systematic reviews has since broadened to include information science ~\citep{kelly_systematic_2013, vakkari_usefulness_2020} and related fields~\citep{vassilakaki_systematic_2015, zawacki-richter_systematic_2019}, effectively informing future developments. The systematic review presents a methodology for establishing valuable findings and distinct contributions in IR beyond a simple overview of past works.


\section{Systematic Review Methodology}

\begin{figure}[!htpb]
\centering
  \includegraphics[width=\linewidth]{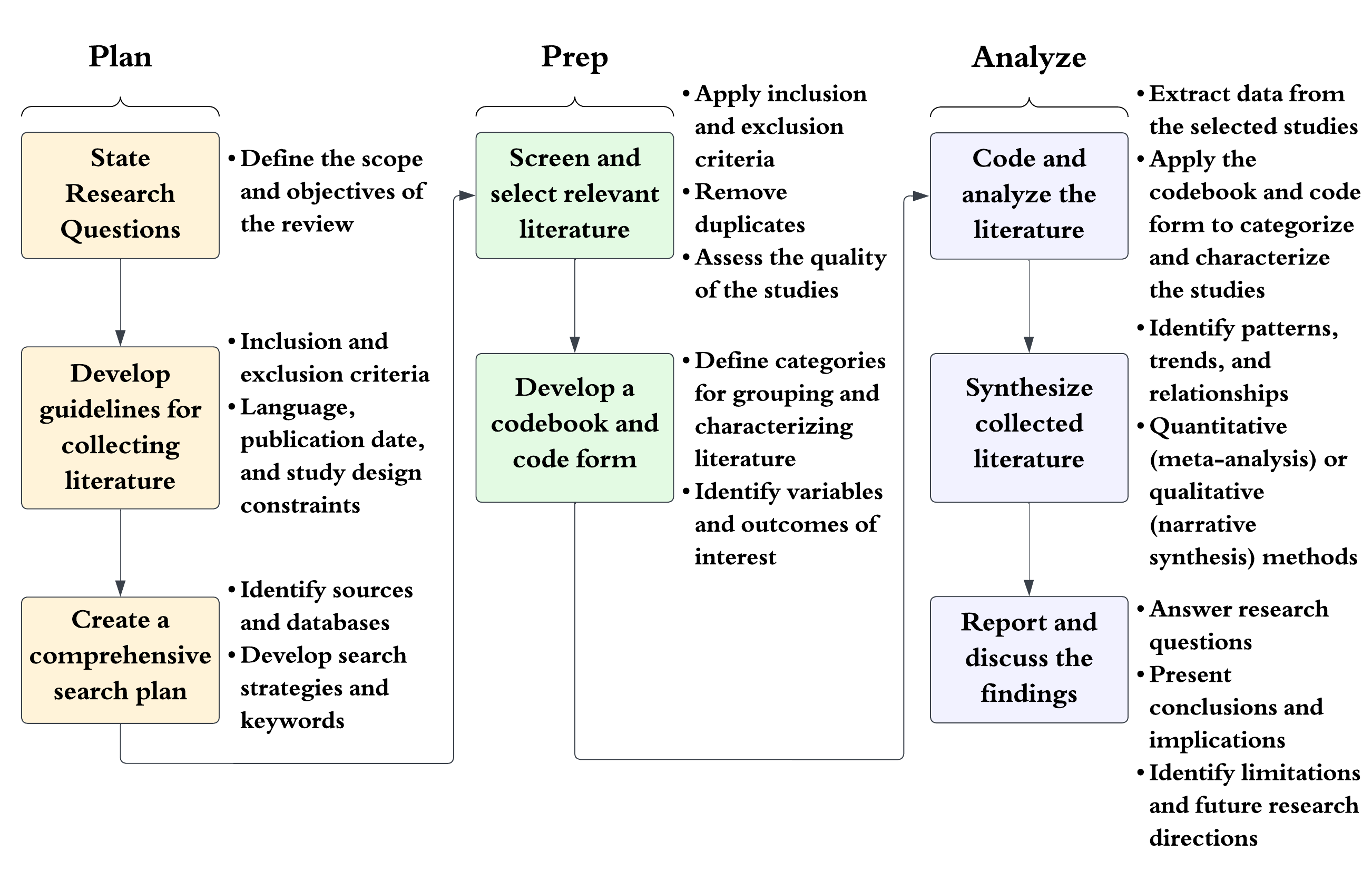}
\caption{{\bf Steps of Performing a Systematic Review.}}
\label{fig1}
\end{figure}

The steps to conducting a systematic review follow three main phases and their sub-components, outlined in 
Figure \ref{fig1}
 and described in detail below. 

\subsection{Selection of Sources (Inclusion/Exclusion Criteria)}

The first step in our systematic review process was the identification of works for study. We focused on literature that directly contributes to the development or capabilities of LLMs. Articles were sourced from Google Scholar and queried using a series of search strings, including: \textit{``large language models,'' ``attention language model,'' 
``generative language models,`` ``natural language understanding,'' ``language model scaling,''  ``prompted language model,'' and ``prompt engineering.''}
This review considered English-language publications from 2016 to 2023.

To identify authoritative sources, we compiled a list of venues known for significant LLM studies, limiting to papers published in select top NLP and AI journals and conferences, as well as reports from leading AI industry developers. The inclusion of industry publications was a deliberate choice aimed at capturing various impactful works in the field, particularly those contributing to novel methodologies and social impact, which might not have been published in academic journals or conferences yet.
In selecting the most influential papers, we applied a citation threshold of 150 citations.
The culmination of this process was the selection of 
61
 articles for our systematic review.

\subsection{Thematic Analysis Procedure / Coding Scheme }

The collected papers were coded for statistical analysis on various categories. Basic metadata features such as authors, publication year, publication venue, and citations were included. We also extracted author affiliations at the time of publication as indicated by the authors’ email domains.

\begin{figure}[!htpb]
\centering
  \includegraphics[width=\linewidth]{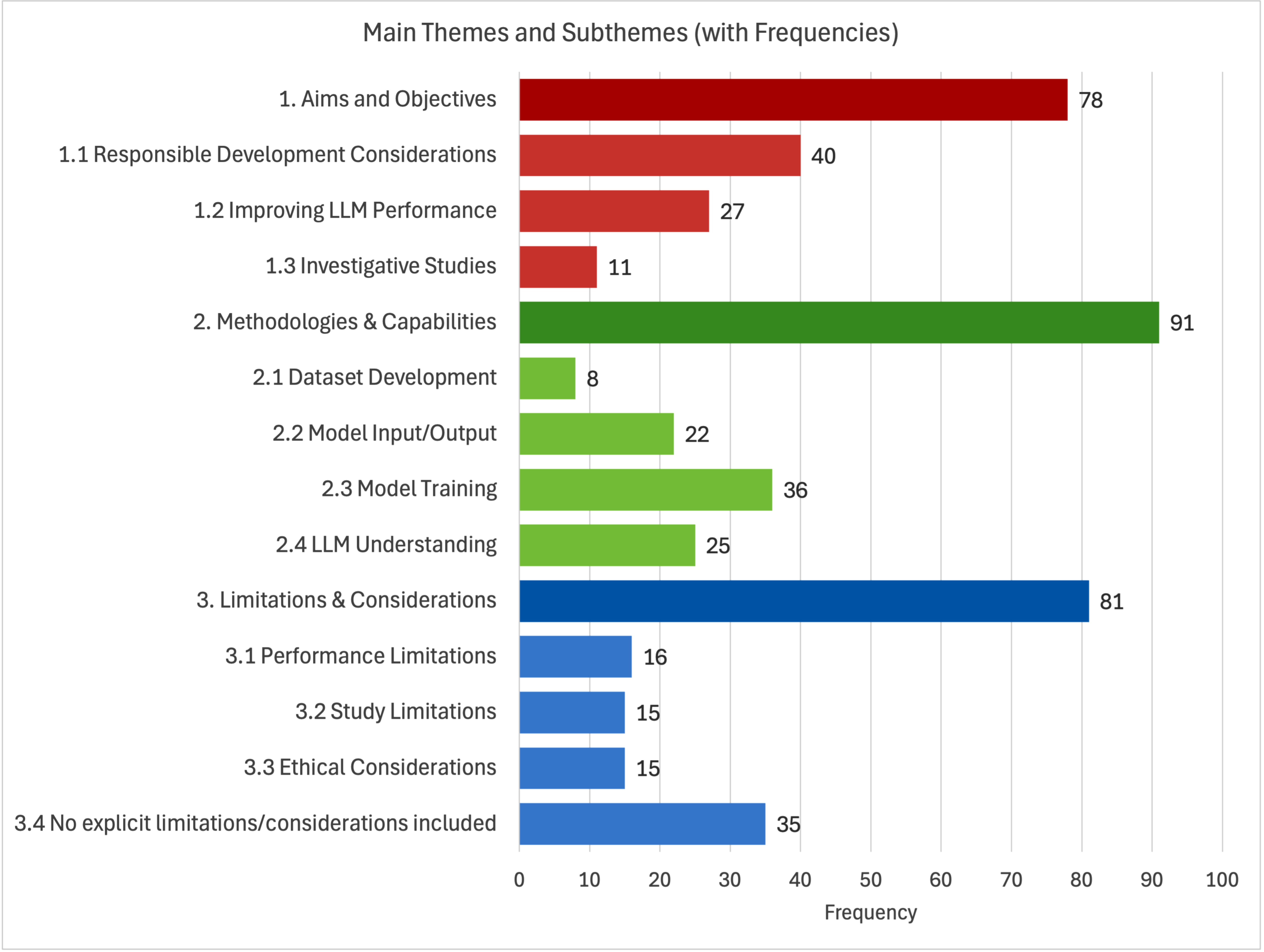}
\caption{\bf Frequencies of Themes and Sub Themes.}
\label{fig2}
\end{figure}

Thematic analysis was performed for the papers across various categories. After reviewing the paper, the authors first produced a list of themes for each piece derived from the qualitative thematic categories. The list was consolidated to generate a coding hierarchy. Each author conducted these steps independently to minimize bias in the process. Then, they compared and discussed the annotations to reach a consensus and resolve any disagreement. 
Figure~\ref{fig2}
shows the distribution of major themes and subthemes for the purpose, methodology, and limitations of the reviewed papers. 
We discuss each of the super themes in individual sections, and the subthemes in corresponding subsections.
Subthemes (e.g., 1.2 Improving LLM Performance) fall under each of these superthemes (e.g., 1. Aims and Objectives), which are shown in the figure. There exists a many-to-many relationship between the articles and the themes, and the themes are not mutually exclusive, as one article can have multiple themes. 

\section{Characteristics of Publications}

In the following section, we provide an overview of the selected articles reviewed, discussing relevant statistics and features that characterize the content of the review.

\begin{figure}[!htpb]
  \centering
  \includegraphics[width=\linewidth]{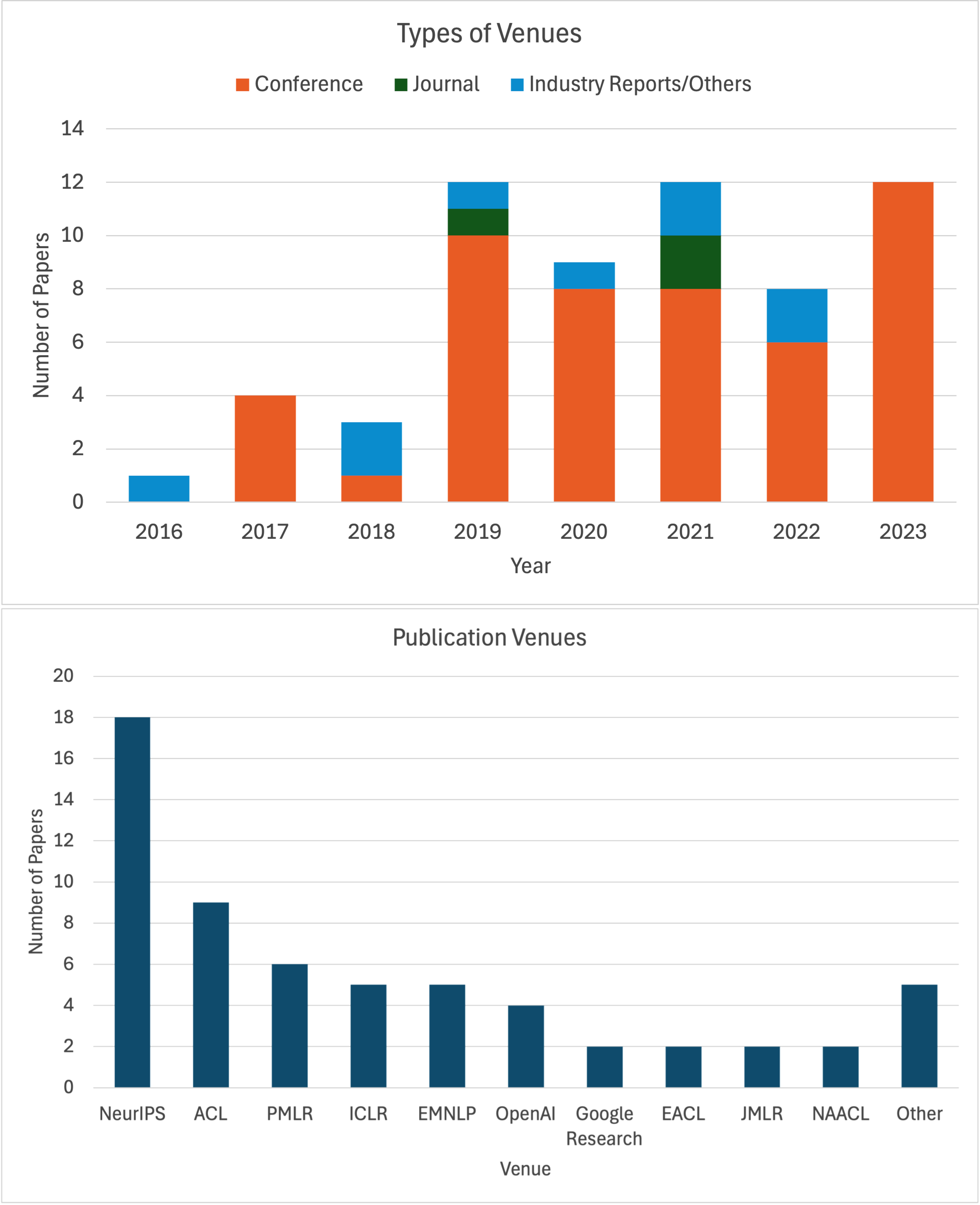}
  \caption{{\bf Publication Statistics.}
  }
  \label{fig3}
\end{figure}

\begin{figure}[!htpb]
  \centering
  \includegraphics[width=0.9\linewidth]{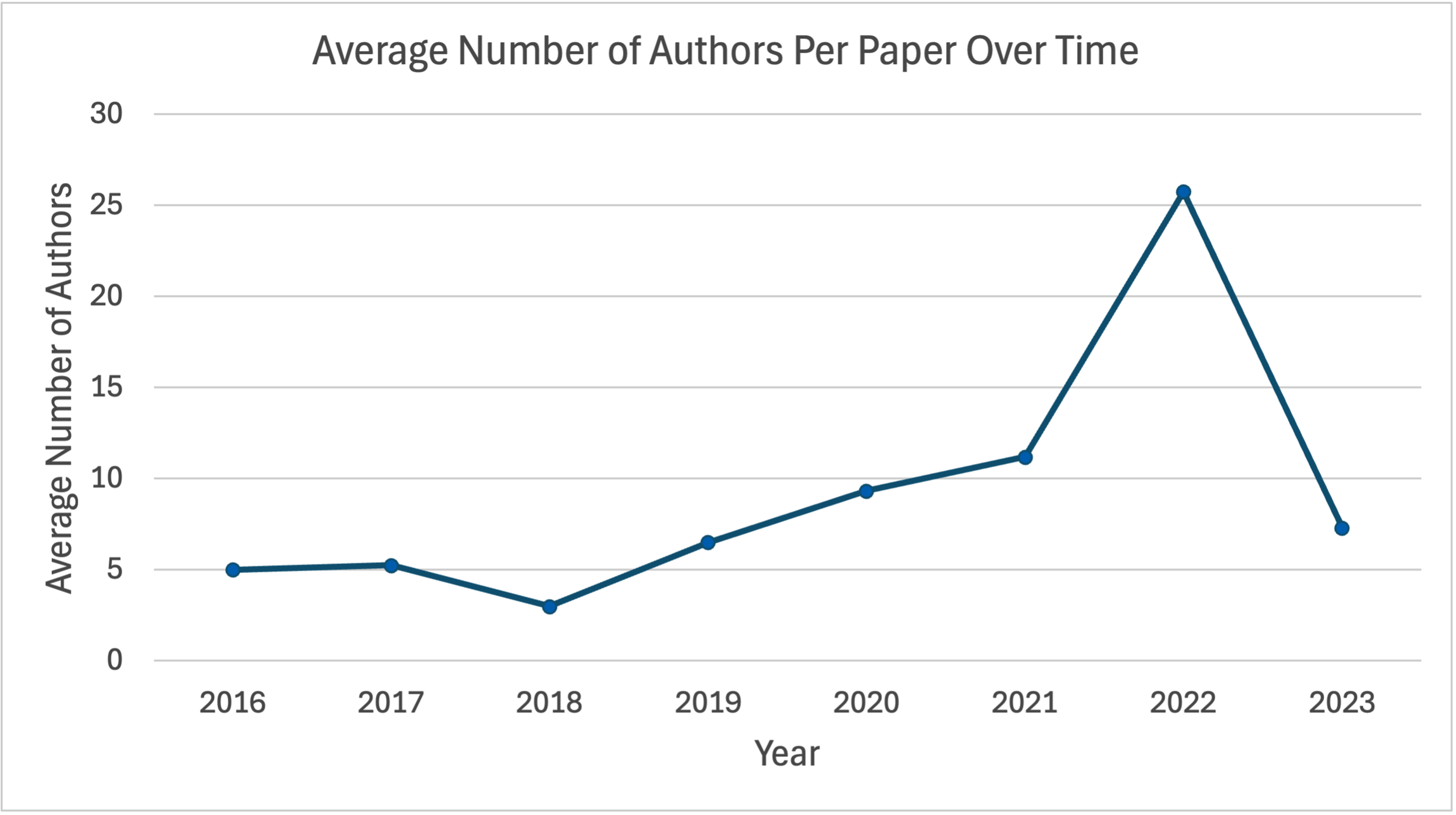}
  \caption{{\bf Avg. Number of Authors Per Paper by Year.}
  }
  \label{fig4}
\end{figure}

Approximately 20\% (n=12) of the articles were published in 2023; 13\% (n=8) in 2022, 20\% (n=12) in 2021; 15\% (n=9) in 2020; 21\% (n=13) in 2019; 3\% (n=2) in 2018; 7\% (n=4) in 2017; and a solitary paper in 2016. Notably, there was a rapid increase in publications between 2018 and 2019. Figure~\ref{fig3} illustrates the distribution of articles across years of publication. Figure \ref{fig3} shows the representation of various types of publication venues—conference papers, journal papers, and industry reports. The reviewed articles represent a selection from 16 venues: 10 conferences, 2 journals, and 4 industry reports. NeurIPS, ACL, and PMLR featured the most articles, with 18, 9, and 6 articles, respectively. ICLR and EMNLP had 5 each, and OpenAI had 4. NAACL, JMLR, EACL, and Google Research had 2 each, and the remaining venues had one each. Articles published in NAACL, NeurIPS, and JMLR garnered the greatest number of citations. Notably, OpenAI’s reports, despite not being in a formally peer-reviewed venue, received a substantial number of citations. 

\begin{figure}[!htpb]
  \centering
  \includegraphics[width=0.7\linewidth]{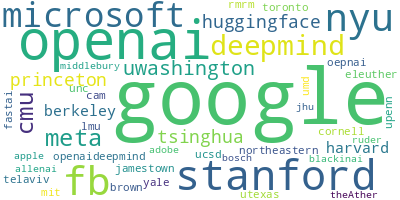}
  \caption{{\bf WordCloud of Author Affiliations.}
  }
  \label{fig5}
\end{figure}

Figure~\ref{fig4} shows the distribution of the number of authors in the articles, and the increase in authorship of top-cited papers over the years. As can be seen in the figure, collaborative works are relatively common among LLM articles; approximately a third (34\%; n=17) of the articles feature over 8 authors, while 4 authors represent the most common size of collaborative groups (18\%; n=11). Figure \ref{fig5} presents the word cloud for organizational affiliations of the authors.
Researchers affiliated with Google, Microsoft, OpenAI, Facebook/Meta, and universities like Stanford, Carnegie Mellon, and New York University are the most represented and have the greatest number of contributors to LLM research. 

\section{Aims \& Objectives (RQ1)}

Literature in the development of LLMs pursues different aims. 
Of the works we studied, a significant 66\% focused on best practices and practical considerations in ethics and research methodology. Meanwhile, 44\% introduced originality in the advancement of LLM performance, focusing on aspects such as efficiency, robustness, and scalability. A smaller proportion, 18\%, 
were investigative studies with the objective of furthering the understanding of LLMs. We explore the main avenues of LLM improvement and progress that are prevalent in the field.

\begin{longtable}{p{2.25cm}p{3.25cm}p{6.25cm}}
\caption{Major Themes for Aims, Objectives, and Purposes of Reviewed Papers} \label{table-purposes} \\
\toprule
Theme & Example Subcategories & Articles \\
\midrule
\endfirsthead

\multicolumn{3}{c}%
{{\bfseries \tablename\ \thetable\ -- continued from previous page}} \\
\toprule
Theme & Example Subcategories & Articles \\
\midrule
\endhead

\midrule
\multicolumn{3}{r}{{Continued on next page}} \\
\endfoot

\bottomrule
\endlastfoot

Responsible Development Considerations &
  Ethics, AI Bias, Information Accuracy, Social Impact, Research Expansion, Responsible Evaluation, Accessibility, Optimal Development, Scaling &
  \citet{lee_deduplicating_2022}; 
  \citet{wang_glue_2019}; 
  \citet{gehman_realtoxicityprompts_2020}; 
  \citet{zhang_opt_2022}; 
  \citet{carlini_extracting_2021}; 
  \citet{lester_power_2021}; 
  \citet{hoffmann_training_2022}; 
  \citet{kaplan_scaling_2020}; 
  \citet{bender_dangers_2021}; 
  \citet{solaiman_release_2019}; 
  \citet{raffel_exploring_2019}; 
  \citet{brown_language_2020}; 
  \citet{fedus_switch_2021}; 
  \citet{rae_scaling_2021}; 
  \citet{vaswani_attention_2017}; 
  \citet{zhang_ernie_2019}; 
  \citet{jiang_how_2021}; 
  \citet{chowdhery_palm_2022}; 
  \citet{lample_cross-lingual_2019}; 
  \citet{wang_superglue_2020}; 
  \citet{kirchenbauer_watermark_2023}; 
  \citet{manakul_selfcheckgpt_2023}; 
  \citet{turpin_language_2023} 
  \\

Improving LLM Performance &
  Data Efficiency, Generalizability, Robustness, Size efficiency, Absolute performance, Diverse language support &
  \citet{sanh_multitask_2022}; 
  \citet{lample_cross-lingual_2019}; 
  \citet{lee_deduplicating_2022}; 
  \citet{wang_glue_2019}; 
  \citet{petroni_language_2019}; 
  \citet{clark_electra_2020}; 
  \citet{dong_unified_2019}; 
  \citet{zhao_calibrate_2021}; 
  \citet{schick_exploiting_2021}; 
  \citet{lester_power_2021}; 
  \citet{howard_universal_2018}; 
  \citet{gururangan_dont_2020}; 
  \citet{radford_improving_2018}; 
  \citet{hoffmann_training_2022}; 
  \citet{jozefowicz_exploring_2016}; 
  \citet{radford_language_2019}; 
  \citet{kaplan_scaling_2020}; 
  \citet{shazeer_outrageously_2017}; 
  \citet{radford_learning_2021}; 
  \citet{raffel_exploring_2019}; 
  \citet{brown_language_2020}; 
  \citet{fedus_switch_2021}; 
  \citet{alayrac_flamingo_2022}; 
  \citet{rae_scaling_2021}; 
  \citet{devlin_bert_2019}; 
  \citet{vaswani_attention_2017}; 
  \citet{liu_multi-task_2019}; 
  \citet{zhang_ernie_2019}; 
  \citet{dauphin_language_2017}; 
  \citet{press_using_2017}; 
  \citet{dai_transformer-xl_2019}; 
  \citet{chowdhery_palm_2022}; 
  \citet{conneau_unsupervised_2020}; 
  \citet{he_deberta_2021}; 
  \citet{wang_superglue_2020}; 
  \citet{yang_xlnet_2019}; 
  \citet{roberts_how_2020}; 
  \citet{kandpal_large_2023}; 
  \citet{zhang_h2o_2023}; 
  \citet{huang_language_2023}; 
  \citet{schick_toolformer_2023}; 
  \citet{yao_tree_2023}; 
  \citet{wang_visionllm_2023}; 
  \citet{gruver_large_2023} 
  \\

Investigative Studies &
  Understanding LLM effectiveness, Understanding LLM knowledge/capabilities &
  \citet{scao_how_2021}; 
  \citet{petroni_language_2019}; 
  \citet{nie_adversarial_2020}; 
  \citet{wei_chain--thought_2022}; 
  \citet{clark_what_2019}; 
  \citet{roberts_how_2020}; 
  \citet{biderman_pythia_2023}; 
  \citet{kandpal_large_2023}; 
  \citet{schaeffer_are_2023}; 
  \citet{turpin_language_2023}; 
  \citet{gruver_large_2023} 
  \\
\end{longtable}

\subsection{Responsible Development Considerations}

As LLMs gain prevalence in social and research contexts, attention has been increasingly directed to the best practices of the field in developing and releasing LLMs~\citep{bender_dangers_2021, solaiman_release_2019, kirchenbauer_watermark_2023, carlini_extracting_2021, gehman_realtoxicityprompts_2020, manakul_selfcheckgpt_2023,turpin_language_2023}.

Various studies have aimed to address the ethical and societal impact of LLMs. For instance, 
\citet{bender_dangers_2021} consider the pervasive risks associated with the growing size of LLMs with a critical review of the limitations and potential dangers of LLMs. 
\citet{solaiman_release_2019} examine the release of LLMs, discussing recommendations for responsible publication in AI that consider the social impacts of OpenAI’s release of the GPT-2 models. Their framework for best practices for responsible model release includes prioritizing and collaborating with the communities influenced by the models rather than focusing on general tradeoffs. 
\citet{kirchenbauer_watermark_2023} 
present a method for `watermarking' LLMs and discuss its applications and security in deployment.

\citet{carlini_extracting_2021} work to demonstrate further the potential for misuse presented by the growth and release of LLMs. They extract memorized individual training points from only black-box query access, showing that LLMs pose a threat to leaking personally identifiable information. 
\citet{gehman_realtoxicityprompts_2020} find biased and harmful propagation learned by LLMs. 
\citet{manakul_selfcheckgpt_2023} and \citet{turpin_language_2023} both note the faultiness of LLMs' information, including their capacity to misinform, hallucinate, and be manipulated.

With these risks in mind, researchers have aimed to implement safe methods in their work and be conscious of the impact of their models. In developing a suite of Open Pre-trained Transformer (OPT) language models, 
\citet{zhang_opt_2022} aim to bring best practices and accessibility in their efficient replication of GPT-3, released with full detailed documentation and an analysis of considerations for the safe deployment of their model.

\subsection{Improving LLM Performance }

Technical advancements are the primary focus of LLM research, and several key works~\citep{radford_language_2019, schick_exploiting_2021, sanh_multitask_2022, brown_language_2020, howard_universal_2018, dong_unified_2019, liu_multi-task_2019, gururangan_dont_2020, wang_glue_2019, gruver_large_2023, zhang_h2o_2023}, 
have concentrated on enhancing the performance of large language models across a diverse range of applications.

A primary objective in LLM research is the generalizability of these models to diverse tasks and domains. For example, 
\citet{radford_language_2019} demonstrate the capacity of LLMs to perform well across various domains and implicitly on downstream tasks without any parameter or architecture modification. Building on the effectiveness of this implicit learning, 
\citet{sanh_multitask_2022} create a model that can better generalize to held-out tasks and perform robustly with diverse prompt wording, using explicit supervised multitask training. 
\citet{brown_language_2020} examine the generalizability of LLMs on new tasks with limited task-specific data in the few-shot setting. 
\citet{gruver_large_2023} examine LLMs zero-shot abilities in the task of time series forecasting. 
\citet{howard_universal_2018} present Universal Language Model Fine-tuning (ULMFiT) –- a transfer learning method –- to pretrain a language model on a large general-domain corpus, applicable to different NLP tasks without task-specific modifications. The GLUE and SuperGLUE benchmarks~\citep{wang_glue_2019, wang_superglue_2020} further this pursuit of generalizability with evaluation and diagnostic datasets spanning diverse tasks and domains.

With the introduction of the BERT architecture, 
\citet{devlin_bert_2019} presented a significant development in language representation models that can be effectively fine-tuned without substantial task-specific architecture modifications. Expanding upon BERT’s capabilities, 
\citet{yang_xlnet_2019} address its limitations with autoregressive language modeling in XLNet.

Many articles target data and size efficiency in LLMs. For instance, 
\citet{lample_cross-lingual_2019} significantly advance effectiveness in tasks pertaining to low-resource languages with limited available data. 
\citet{zhao_calibrate_2021} improve data efficiency by resolving the problem of stability in few-shot learning. 
\citet{roberts_how_2020} leverage LLMs’ ability to implicitly store and retrieve knowledge to determine its utility in answering questions without additional training. To maximize the use of large unlabeled data, 
\citet{radford_improving_2018} employ unsupervised generative pretraining to enhance natural language understanding performance with limited data. The ELECTRA~\citep{clark_electra_2020} and DeBERTa architectures~\citep{he_deberta_2021} improve BERT’s sample-efficiency. 
\citet{kaplan_scaling_2020}
and \citet{zhang_h2o_2023} 
tackle efficiency in model development by optimizing performance based on the factors of compute power. These works collectively target various aspects of model performance to further the effectiveness of LLMs.

\subsection{Investigative Studies}

The objective of some papers is to delve into the less understood mechanisms that drive the success of LLMs~\citep{clark_what_2019, scao_how_2021, wei_chain--thought_2022, turpin_language_2023}. 

Given the complexity yet significant effectiveness of attention mechanisms, 
\citet{clark_what_2019} decipher the patterns in LLM attention and their association with specific linguistic features. In examining the performance gains achieved through prompting, 
\citet{scao_how_2021} analyze the effectiveness of prompting and the trends in its success. 
\citet{wei_chain--thought_2022} experiment with chain-of-thought prompting to gauge LLMs’ reasoning. Inspired by the human tendency to break down complex tasks into multi-step problems, they augmented the input examples with a chain of thought. 
Meanwhile, \citet{turpin_language_2023} studied the weaknesses of chain-of-thought prompting and where it could result in unfaithful responses and be prone to manipulation.

Such investigative studies offer valuable insights into the capacities of LLMs and inform further development. 
\citet{roberts_how_2020} leverage LLMs’ implicit knowledge storage in question-answering tasks. LLMs are able to achieve competitive results in question-answering without access to external resources, offering insights into the uses and underlying learning processes of LLMs. Similarly, 
\citet{petroni_language_2019} explore the potential of LLMs as knowledge bases, building on this inherent characteristic. 
\citet{nie_adversarial_2020} introduce a benchmark, Adversarial NLI, to identify and evaluate tasks that pose the greatest challenge for LLMs, like numerical and quantitative reasoning and complex inference types. These works all aim to better empirically understand LLMs. 
\citet{biderman_pythia_2023} present Pythia, which enables the analysis of LLM capabilities with scale and training. \citet{schaeffer_are_2023} argue that the unique abilities of larger-scale models may be a `mirage' resulting from the metrics used based on their mathematical model examining changes in LLM performance with size.


\section{Methodologies \& Capabilities (RQ2)}

Various stages of model development contribute significantly to the efficacy of LLMs, presenting multiple facets to address in refining their capabilities. 
In our review, 13\% of the works' methodologies involved the development of datasets specifically for LLMs. 36\% focused on studying model inputs and outputs, including aspects such as prompting, formatting, and pre-/post- processing techniques. The majority, 59\%, delved into model training, examining a spectrum of architectures. A significant proportion, 41\%, centered their methods on the analysis of LLMs.
Through a comprehensive analysis of the methods employed, we aim to provide valuable insights into the key strategies and their contributions, illuminating paths for continued improvement and innovation.

\begin{longtable}{p{2.25cm}p{3.25cm}p{6.25cm}}
\caption{Major Themes for Methodologies and Capabilities of Reviewed Papers} \label{table-methods} \\
\toprule
Theme & Example Subcategories & Articles \\ \midrule
\endfirsthead

\multicolumn{3}{c}%
{{\bfseries Table \thetable\ continued from previous page}} \\
\toprule
Theme & Example Subcategories & Articles \\ \midrule
\endhead

\midrule
\multicolumn{3}{r}{{Continued on next page}} \\
\endfoot

\bottomrule
\endlastfoot

  Dataset Development 
  &
  Diverse Datasets, Challenging Datasets, Dataset Construction Methods, Benchmarks 
  &
  \citet{gehman_realtoxicityprompts_2020}; \citet{wang_glue_2019}; 
  \citet{wang_superglue_2020}; 
  \citet{radford_language_2019}; 
  \citet{petroni_language_2019}; 
  \citet{nie_adversarial_2020}; 
  \citet{huang_language_2023}; \citet{manakul_selfcheckgpt_2023}
  \\

  Model Input/Output 
  &
  Task Format, Prompting, Text-to-Text, Task-specific Input, Character-Level, Tokens, Data Preprocessing, Decoding, Output Calibration 
  &
  \citet{sanh_multitask_2022}; 
  \citet{scao_how_2021}; 
  \citet{lee_deduplicating_2022}; \citet{gehman_realtoxicityprompts_2020}; \citet{zhao_calibrate_2021}; 
  \citet{schick_exploiting_2021}; 
  \citet{lester_power_2021}; 
  \citet{gururangan_dont_2020}; 
  \citet{raffel_exploring_2019}; \citet{vaswani_attention_2017}; 
  \citet{press_using_2017}; 
  \citet{wei_chain--thought_2022}; 
  \citet{jiang_how_2021}; 
  \citet{gao_making_2021};
  \citet{kirchenbauer_watermark_2023};
  \citet{kandpal_large_2023}; 
  \citet{zhang_h2o_2023}; 
  \citet{huang1}; 
  \citet{schick_toolformer_2023}; 
  \citet{yao_tree_2023}; 
  \citet{wang_visionllm_2023}; 
  \citet{gruver_large_2023} 
  \\

  Model Training
  &
  Cross-lingual learning,
Multitask learning, Unsupervised learning, Fine-tuning methods, Model architecture, Training Objectives, Multimodal, Scaling 
&

  \citet{sanh_multitask_2022}; 
  \citet{lample_cross-lingual_2019}; 
  \citet{scao_how_2021}; \citet{gehman_realtoxicityprompts_2020}; \citet{petroni_language_2019}; 
  \citet{zhang_opt_2022}; 
  \citet{clark_electra_2020}; 
  \citet{dong_unified_2019}; 
  \citet{schick_exploiting_2021}; 
  \citet{lester_power_2021}; 
  \citet{howard_universal_2018}; 
  \citet{gururangan_dont_2020}; \citet{radford_improving_2018}; \citet{hoffmann_training_2022}; \citet{jozefowicz_exploring_2016}; \citet{radford_language_2019}; 
  \citet{kaplan_scaling_2020}; \citet{shazeer_outrageously_2017}; \citet{radford_learning_2021}; \citet{raffel_exploring_2019}; 
  \citet{brown_language_2020}; 
  \citet{fedus_switch_2021}; 
  \citet{alayrac_flamingo_2022}; 
  \citet{rae_scaling_2021}; 
  \citet{devlin_bert_2019}; 
  \citet{vaswani_attention_2017}; 
  \citet{liu_multi-task_2019}; 
  \citet{zhang_ernie_2019}; 
  \citet{dauphin_language_2017}; 
  \citet{dai_transformer-xl_2019}; \citet{chowdhery_palm_2022}; 
  \citet{gao_making_2021}; 
  \citet{conneau_unsupervised_2020}; 
  \citet{he_deberta_2021}; 
  \citet{yang_xlnet_2019}; 
  \citet{saharia_photorealistic_2022}
   \\

LLM Understanding &
  Factor analysis, Metrics, Optimal practices, Risks, Review &

  \citet{scao_how_2021}; 
  \citet{wang_glue_2019}; \citet{gehman_realtoxicityprompts_2020}; \citet{carlini_extracting_2021}; 
  \citet{dong_unified_2019}; 
  \citet{zhao_calibrate_2021}; 
  \citet{gururangan_dont_2020}; \citet{hoffmann_training_2022}; 
  \citet{kaplan_scaling_2020}; 
  \citet{bender_dangers_2021}; 
  \citet{radford_language_2019}; \citet{solaiman_release_2019}; \citet{raffel_exploring_2019}; 
  \citet{brown_language_2020}; 
  \citet{vaswani_attention_2017}; 
  \citet{dai_transformer-xl_2019}; 
  \citet{wei_chain--thought_2022}; 
  \citet{jiang_how_2021}; 
  \citet{clark_what_2019}; 
  \citet{conneau_unsupervised_2020}; 
  \citet{he_deberta_2021};
  \citet{roberts_how_2020}; 
  \citet{turpin_language_2023}; 
  \citet{schaeffer_are_2023}; 
  \citet{kandpal_large_2023}; 
  \citet{biderman_pythia_2023} 
  \\
\end{longtable}

\subsection{Dataset and Benchmark Development}

Large, high-quality corpora are foundational to the capabilities of LLMs. The development of diverse and effective datasets has been a pivotal method in furthering LLMs, and our review identified several studies that explored new datasets~\citep{wang_glue_2019, wang_superglue_2020, nie_adversarial_2020, radford_language_2019, huang_language_2023,manakul_selfcheckgpt_2023}.

The General Language Understanding Evaluation (GLUE) benchmark~\citep{wang_glue_2019} has emerged as one of the most significant datasets in NLP, widely used to evaluate models' NLU capacity. This work presents a dataset comprising diverse linguistic tasks and domains to advance models' sample-efficient knowledge transferability and a diagnostic evaluation dataset tagged with various linguistic phenomena to facilitate error analysis. With the rapid advancement of LLMs surpassing human-level performance, the authors later introduced SuperGLUE~\citep{wang_superglue_2020} to further challenge LLMs with more demanding tasks, varied task formats, and comprehensive human baselines. 

Continuing the pursuit of more demanding datasets to facilitate LLM growth, 
\citet{nie_adversarial_2020} propose an adversarial approach to constructing a dynamic benchmark for longevity, Adversarial NLI. In this approach, human annotators iteratively create intentionally challenging examples to pinpoint and expose model weaknesses.

\citet{radford_language_2019} investigate the capacity of LLMs to learn implicitly from data by constructing a large and manually filtered corpus, WebText, aimed at training a model on varied domains and contexts for improved applicability across a broader range of tasks. The data underpinning LLMs is crucial to their linguistic and relational knowledge. As such, they can inadvertently learn toxicity from texts and generate biased content. 
\citet{gehman_realtoxicityprompts_2020} address the risks posed by flawed datasets by constructing REALTOXICITYPROMPTS, a dataset of naturally occurring prompts with toxicity scores, quantifying the risk of a pretrained language model for generating toxic text. 
\citet{manakul_selfcheckgpt_2023} develop SelfCheckGPT, a method that enables fact-checking of LLMs' hallucinations. \citet{huang_language_2023} develop a multimodal text-image mixed corpus to train Kosmos-1 a powerful multimodal LLM. 
Dataset development has emerged as a prevalent method to both expand and rigorously test LLM capabilities.

\subsection{Model Input/Output}

The formatting and processing of natural language data for use in models is a relevant aspect of LLM development that has seen new methodologies~\citep{scao_how_2021, gao_making_2021, sanh_multitask_2022, wei_chain--thought_2022, lester_power_2021, wang_visionllm_2023, schick_toolformer_2023}.

The interpretation of tasks as prompts has arisen as an effective paradigm for adapting pretrained models. Scao \& Rush (2021)~
\citet{scao_how_2021} establish the data efficiency provided by prompting and propose a novel metric to quantify the advantage of a prompt over a generic model head across tasks and data sizes. Given the benefits of prompts, literature has shifted towards improving and optimizing prompt generation. Gao et al. (2021)~
\citet{gao_making_2021} present LM-BFF, incorporating effective techniques to test the limits of the prompting methodology. 
\citet{wang_visionllm_2023} approach images as a foreign language and aligning vision tasks with language tasks through the input format of natural language instructions. 

Delving into the low-data applications of prompting, 
\citet{sanh_multitask_2022} demonstrate the capacity of prompting for implicit multitask learning, enhancing zero-shot generalization. 
\citet{wei_chain--thought_2022} leverage processes in human problem-solving to examine ``chains of thought'' in prompting to improve reasoning-based tasks in LLMs. Adapting prompting with fine-tuning, 
\citet{lester_power_2021} introduce ``soft prompts,'' which are learned end-to-end through backpropagation from a pretrained model and incorporated into downstream tasks. 
\citet{yao_tree_2023} further presents a `tree of thoughts' to improve LLMs' abilities in problem-solving tasks. 

Many studies have explored the modification of inputs with new formats or additional information to support model learning~\citep{raffel_exploring_2019, vaswani_attention_2017, press_using_2017, gehman_realtoxicityprompts_2020}. 
\citet{raffel_exploring_2019} introduced an influential work with a unified framework that converts text-based tasks into a text-to-text format. 
\citet{schick_toolformer_2023} introduce APIs into LLMs, allowing their model Toolformer to access and use external tools to support their responses. 
\citet{schick_exploiting_2021}, motivated by including natural language task descriptions in inputs, present Pattern-Exploiting Training (PET), which reformulates inputs as cloze-style phrases. 
\citet{vaswani_attention_2017}, in their Transformer architecture, add positional encodings to the input embeddings, enhancing efficiency. 
\citet{press_using_2017} alter input embeddings by tying the input embedding to the output embedding. 
\citet{gehman_realtoxicityprompts_2020} adopt an approach of inserting information in inputs to target their limitations. In particular, they use prefix tokens to reduce toxicity in generations.

Outputs are also areas of development~\citep{gehman_realtoxicityprompts_2020, zhao_calibrate_2021}. In examining ways to address model toxicity, 
\citet{gehman_realtoxicityprompts_2020} present decoding-based intervention methods. 
\citet{zhao_calibrate_2021} implement calibration of outputs based on generations to neutralize inputs to improve few-shot bias.

\subsection{Model Training}

New methods of model architecture development bring significant advancements to the capabilities of LLMs.
In particular, developments in the pretraining processes and objectives are a prevalent area of study~\citep{devlin_bert_2019, clark_electra_2020, yang_xlnet_2019, radford_improving_2018, dong_unified_2019, lample_cross-lingual_2019}. 

For large-scale language modeling with RNNs, 
\citet{jozefowicz_exploring_2016} introduce a Softmax loss based on character-level CNNs and combine word- and character-level models, integrating them into the word-level LSTM. 
\citet{radford_learning_2021} explore pretraining methods for CNNs in multimodal learning with vision-based tasks, introducing Contrastive Language-Image Pre-training (CLIP). This method involves pretraining by learning a joint multimodal embedding with an image and text encoder and then predicting image-text pairings.

\citet{vaswani_attention_2017} introduce the Transformer architecture, featuring the self-attention mechanism to learn the relationships between words within sentences, enabling greater scale and efficiency.
The development of BERT~\citep{devlin_bert_2019} was one of the most important advances of LLMs. BERT is designed to pretrain generalizable bidirectional representations from unlabeled text. Its pretraining employs two unsupervised tasks: Masked Language Modeling (MLM), where masked tokens are predicted, and Next Sentence Prediction (NSP), where the relationship between two sentences is predicted. Aiming for sample efficiency and reduced computational cost, 
\citet{clark_electra_2020} propose a novel pretraining method through ELECTRA. In this method tokens are replaced with alternatives sampled from a small generator and the model learns to discriminate between generated and true tokens. 
\citet{yang_xlnet_2019} address the limitations of BERT’s autoencoding-based pretraining in learning masked dependencies with their autoregressive pretraining method in XLNet. 
\citet{radford_improving_2018} use generative pretraining – an innovative approach to pretraining – for LLMs, followed by discriminative fine-tuning for specific tasks. ERNIE~\citep{zhang_ernie_2019} uses knowledge graphs as input to provide structured information alongside natural language.
\citet{dong_unified_2019} introduce a new unified pre-training language model (UNILM), jointly optimized for multiple objectives to enable fine-tuning for both NLU and NLG tasks. This approach enhances task generalizability. To broaden capabilities across languages, 
\citet{lample_cross-lingual_2019} introduce two learning methods for cross-lingual language modeling.

\subsection{LLM Understanding}

Various research works attempt to advance LLMs through analysis-based methods~\citep{scao_how_2021, gehman_realtoxicityprompts_2020, zhao_calibrate_2021, jiang_how_2021}.

The development of metrics to quantify aspects of LLMs enhances understanding of both the limitations and effectiveness of these models. 
\citet{scao_how_2021} introduce an approach to quantify the advantage of prompting, aiming to guide practices in developing LLMs with scale by examining the impact of prompting across various data sizes. 
\citet{gehman_realtoxicityprompts_2020} REALTOXICITYPROMPTS dataset provides valuable insights for gauging the risks of LLMs and understanding the factors to be conscious of. Metrics such as Majority Label Bias, Recency Bias, and Common Token Bias are crucial~\citep{zhao_calibrate_2021} to characterize model stability, aiding in the development of their calibration approach to address instability. 
\citet{jiang_how_2021} propose a set of methods to better measure the accuracy of model knowledge that accounts for the role of prompts in the quality of generations. Their work enables a more accurate estimation of the knowledge in language models with the automatic generation of better prompts.


\section{Limitations \& Considerations (RQ3)}

Understanding the limitations of studies and LLMs provides essential context and guidance for future work. Of the articles we examined, 43\% explicitly recognized the limitations and ethical considerations of their study. The proportion of articles including these limitations has increased with recency, as the importance of discussing limitations has become increasingly emphasized, with some venues even requiring specific sections for them. Across these papers, 62\% noted limitations regarding performance, 58\% identified limitations in the study procedure, and 58\% examined the ethical impacts of their work. In our analysis, we examine the acknowledged limitations in the development of LLMs to promote best practices in LLM research.

\begin{longtable}{p{2.25cm}p{3.25cm}p{6.25cm}}
\caption{Limitations and Considerations of LLM Research} \label{table-limitations} \\
\toprule
Theme & Example Subcategories & Articles \\ \midrule
\endfirsthead

\multicolumn{3}{c}%
{{\tablename\ \thetable\ -- continued from previous page}} \\
\toprule
Theme & Example Subcategories & Articles \\ \midrule
\endhead

\midrule
\multicolumn{3}{r}{{Continued on next page}} \\
\endfoot

\bottomrule
\endlastfoot

Performance Limitations &
  Low-data settings, Complex tasks, Generalizability, Interpretability &
  \citet{sanh_multitask_2022}; 
  \citet{lee_deduplicating_2022}; 
  \citet{zhao_calibrate_2021}; 
  \citet{lester_power_2021}; 
  \citet{kaplan_scaling_2020}; 
  \citet{radford_language_2019}; 
  \citet{radford_learning_2021}; 
  \citet{rae_scaling_2021}; 
  \citet{wang_glue_2019};
  \citet{biderman_pythia_2023}; \citet{kirchenbauer_watermark_2023}; \citet{manakul_selfcheckgpt_2023}; \citet{huang_language_2023}; 
  \citet{schick_toolformer_2023}, 
  \citet{yao_tree_2023}; 
  \citet{gruver_large_2023}
  \\

Study Limitations &
  Practical Replicability Limitations, Cost, Compute power, Limited Study Scope, Imperfect assumptions, Imperfect Evaluation &
  \citet{sanh_multitask_2022}; 
  \citet{scao_how_2021}; 
  \citet{lee_deduplicating_2022}; 
  \citet{wang_glue_2019}; 
  \citet{gehman_realtoxicityprompts_2020}; 
  \citet{dong_unified_2019}; 
  \citet{zhao_calibrate_2021}; 
  \citet{kaplan_scaling_2020}; 
  \citet{bender_dangers_2021}; 
  \citet{rae_scaling_2021}; 
  \citet{radford_learning_2021};
  \citet{kandpal_large_2023}; 
  \citet{schaeffer_are_2023}; 
  \citet{turpin_language_2023}; 
  \citet{yao_tree_2023}
  \\

Ethical Considerations &
  Harmful generation, Model release risks, Environmental impact, Language limitations, Language limitations &
  \citet{sanh_multitask_2022}; 
  \citet{scao_how_2021}; 
  \citet{gehman_realtoxicityprompts_2020}; 
  \citet{dong_unified_2019}; 
  \citet{lester_power_2021}; 
  \citet{bender_dangers_2021}; 
  \citet{radford_learning_2021}; 
  \citet{alayrac_flamingo_2022}; 
  \citet{gao_making_2021}; 
  \citet{wang_glue_2019}; 
  \citet{biderman_pythia_2023}; 
  \citet{kandpal_large_2023}; 
  \citet{huang_language_2023}; 
  \citet{yao_tree_2023}; 
  \citet{wang_visionllm_2023}  
  \\
  
\end{longtable}

\subsection{Performance Limitations}

A significant and widely recognized limitation in LLM research is their weaknesses in performance for robust applicability. The articles have identified the limited capacity of LLMs in tackling certain complex tasks~\citep{wang_glue_2019, radford_language_2019, brown_language_2020, gao_making_2021, solaiman_release_2019, schick_toolformer_2023}. 
\citet{wang_glue_2019}, in examining the performance of NLU systems through their construction of the GLUE benchmark and its diagnostic dataset, note the challenges LLMs face on specific tasks and various linguistic phenomena: models often perform lower on the RTE and WNLI inference tasks, and Logic-based tasks, and struggle in cases involving restrictivity and double negation.

\citet{radford_language_2019} recognize that in tasks such as summarization, LLM performance is lower, a finding echoed by 
\citet{brown_language_2020}, who identify tasks like text synthesis as a weakness. The performance disparities in harder tasks are noted by 
\citet{gao_making_2021}, who observe that their method favors tasks with shorter inputs, fewer output classes, and straightforward structures amenable to a ``fill-in-the-blank'' format. 
\citet{solaiman_release_2019} further note long input text as a challenge in the accuracy of LLMs.
The growing capabilities of LLMs have relied heavily on the availability of large amounts of data, leaving low-data and zero- to few-shot settings as a persisting challenge of LLMs~\citep{brown_language_2020, alayrac_flamingo_2022, radford_learning_2021, sanh_multitask_2022, gao_making_2021}. 
\citet{brown_language_2020}, 
\citet{schick_toolformer_2023}, 
and 
\citet{alayrac_flamingo_2022} both note poor sample efficiency in training as a broader limitation of LLMs. In their analyses, 
\citet{radford_language_2019} characterize the zero-shot performance of GPT-2 as practically ``far from usable,'' mirroring the substantial gap in zero-shot performance from traditionally fine-tuned models identified by 
\citet{sanh_multitask_2022}. Beyond absolute performance, 
\citet{gao_making_2021} also identify substantial instability in few-shot learning. LLMs experience limited broad generalizability. 
\citet{biderman_pythia_2023} analyze the capabilities of LLMs with scale using their suite of models Pythia, showing that with growing scale, LLMs improve in performance, albeit with a decreasing rate. The low zero-shot performance of LLMs may be less relevant for other and newer LLMs, but is still prominent. These present an important and relevant area for further study.

\subsection{Study Limitations}

Beyond the technical limitations of LLMs, the articles examine the limitations of their study approaches and analyses. To maintain best practices and provide context for their findings, multiple articles acknowledge the limitations of their study scope and its imperfect assumptions~\citep{gehman_realtoxicityprompts_2020, hoffmann_training_2022, chowdhery_palm_2022, roberts_how_2020, scao_how_2021, lee_deduplicating_2022, radford_learning_2021, schaeffer_are_2023}.

The toxicity dataset composed by 
\citet{gehman_realtoxicityprompts_2020} relies on a single available metric of toxicity detection in text, which they acknowledge as an imperfect measure that could bias toxicity detection towards lexical cues and miss subtle biases. 
\citet{hoffmann_training_2022}, 
\citet{turpin_language_2023}, and \citet{manakul_selfcheckgpt_2023} 
similarly note their limited scope and scale of study as a constraint to the broader generalizability of their findings.

\citet{roberts_how_2020} propose analyses beyond their scope, including evaluating tasks that require reasoning capabilities, for future work. 
\citet{schaeffer_are_2023} also recognize their limited scope. 
In addition to expanding on the limitations of their evaluation datasets, 
\citet{radford_learning_2021} recognize their utilization of validation sets that do not accurately reflect true zero-shot scenarios despite their zero-shot focus as a limitation of their work in broader applicability for low-data settings.
\citet{scao_how_2021} acknowledge their use of human-written prompts and recommend the analysis of automatic prompts for future work. 
\citet{lee_deduplicating_2022} clarify the focuses of their study, describing that they examine the over-representation of duplicate texts but not under-representation, and they do not distinguish or analyze the positive versus negative impact of memorized content. These limited scopes introduce room for further examinations and analyses, as well as additional implications about the context of the findings presented.

\subsection{Societal Impact}

The limitations of language models as they pertain to large-scale release, application, and social impact are especially important to consider as LLMs grow in size and reach. 

The potentially harmful effects of LLMs pose risks in model release~\citep{alayrac_flamingo_2022, radford_learning_2021, brown_language_2020, saharia_photorealistic_2022, carlini_extracting_2021, zhang_opt_2022, sanh_multitask_2022, yao_tree_2023, huang_language_2023}. The most prominent mention of these risks is in their bias and toxicity, which are problems common across the use of LLMs, noted in the multimodal work of
\citet{alayrac_flamingo_2022} and 
\citet{radford_learning_2021} and the scaling studies of 
\citet{rae_scaling_2021} and 
\citet{hoffmann_training_2022}. 
In addition to issues of bias, 
\citet{brown_language_2020}, 
\citet{saharia_photorealistic_2022}, 
\citet{yao_tree_2023}, \citet{huang_language_2023}, 
and 
\citet{carlini_extracting_2021} highlight the potential for deliberate misuse of LLMs and their privacy limitations.

These concerns about bias and harmful uses bring risks in the release of LLMs, necessitating a balance between security and openness. 
\citet{zhang_opt_2022} cite these reasons as justification for releasing their OPT models to the research community before broader deployment. 
\citet{sanh_multitask_2022} emphasize transparency and reproducibility as guiding principles in their decision to release their dataset, models, and tools. However, they acknowledge that their deliberate decisions to reduce the use of corpora with harmful content do not fully eliminate biases in LLMs.

Various articles also acknowledge the limitations of LLMs’ benefits in terms of their environmental impact 
\citep{zhang_opt_2022,scao_how_2021,sanh_multitask_2022,bender_dangers_2021}. 
\citet{wang_visionllm_2023} make efforts to reduce training resources and thus the carbon footprint of their model. 
While LLMs are powerful tools, they can have limitations in their readiness and ability to make a positive impact on society. This underscores the need for a balanced approach to assessing their benefits and drawbacks.


\section{Discussion}

In this section, we answer the research questions and discuss the implications and future directions in LLM research.

\subsection{Answering Research Questions}

To answer RQ1, we examined the main objectives of LLM research to identify the most prevalent areas to address in the field. One primary aim was to improve the absolute performance of LLMs, focusing on their efficiency, robustness, and scalability. With the increasing applications of LLMs, many studies also aimed to develop and apply best practices from both a technical and an ethical standpoint. Others bridged these two areas, analyzing both the abilities and impacts of LLMs.

In RQ2, we explored the methodologies of LLM studies to understand the various approaches to advancing LLM capabilities. The results demonstrated a focus on every stage of LLM development. A significant portion of the studies targeted model architectures, developing new learning strategies, and training and pre-training objectives. Concurrently, other studies highlighted the role of factors external to the model, such as constructing new datasets and processing and interpreting inputs and outputs. A set of studies applied methods that analyzed LLMs’ capabilities.

To answer RQ3, we thoroughly examined the LLM studies and their ethical considerations. Firstly, we found weaknesses in the performance of LLMs on complex tasks, such as those involving logic, longer text, and specific linguistic phenomena. Furthermore, the data-reliance of LLMs raises concerns about their robustness. Secondly, we identified areas for expansion in the limited scope and assumptions used in the studies. Lastly, we found important considerations regarding the impact of LLMs, including learned toxicity and biases, potential for misuse, and additional factors that hold significant implications for the release of LLMs.

\subsection{Implications and Future Work}

Beyond summarizing current trends in LLM research, this systematic review has pinpointed pivotal directions and key considerations for future studies.

\subsubsection{Research Topics}

We identify gaps and areas of progress in current research. In particular, the most underaddressed yet valuable areas include the scaling of LLMs, enhancing LLMs’ linguistic understanding and reasoning capabilities, and improving the data efficiency of LLMs. Furthermore, as LLMs evolve and their capabilities expand, understanding how they learn and what drives their success becomes increasingly crucial to inform further development. Therefore, analysis and investigations into LLM explainability and interpretability are important.

The originality of a study is characterized by the combination of its purpose and methods: we reviewed works that aim to improve generalizability through model architecture, address bias in post-processing, or enhance efficiency with optimized parameters. These insights pave the way for future work to be informed by these approaches, advancing each component of LLM development and contributing to the collective progress of LLMs as a whole.

\subsubsection{Responsible Development}

The studies we examined highlighted the growing emphasis on measures to address the risks and societal impacts of LLM research. A primary focus is on extensive documentation in dataset and model development, which is valuable not only for replicability but also for transparency and awareness of both the ethical and technical limitations. In-depth documentation, including examination and measurement of potential harmful risks and biases, has been a focal point across the objectives, methodologies, and limitations of LLM research.

Additionally, we underscore the need for cautious considerations in how LLMs are made available to the community and the importance of collaboration in the development and release of LLMs. Incorporating discussions of the limitations and impacts of studies is becoming an essential component of research. Furthermore, we examine studies that present considerations for more effective and efficient development, such as optimal processing, parameters, and architectures.

\subsection{Limitations}

Although this systematic review aims to present a comprehensive perspective on examining LLM research and the impacts of NLP, there are several limitations. First, while the literature search endeavored to include as diverse a range of articles as possible, the possibility of missing relevant studies exists. 
While we experimented with several repositories for scholarly literature in the early phases of this study, we used Google Scholar to identify the final set of papers and to record the numbers for reporting purposes. Google Scholar could be biased toward specific articles, and we should have cross-referenced the returned results with other bibliometric databases to identify a fairer collection of scholarly articles.
To mitigate this in future research, more comprehensive queries that encompass a broader array of articles could be employed, and a wider range of publication venues and bibliometric databases could be included. Another limitation lies in the annotation of the themes of each paper. While we exercised due diligence in ensuring that categories and groupings were generated to best represent the articles, with discussions following each analytical phase, involving a larger number of reviewers could enhance the reliability, consistency, and replicability of these qualitative themes.


\section{Conclusion}

LLMs are among the most influential developments in the field of NLP and AI, with rapid advancements making significant technical and societal impacts. However, the capabilities of LLMs are still rapidly evolving. Our aim was to examine the advancements and impacts of LLMs through a systematic review of studies on LLMs. The results contribute to a better understanding of the capabilities of LLMs and LLM research methodologies, the goals and avenues of the field, and the research and ethical limitations in LLMs. We also emphasize the value of studies that aim to understand and explain the workings of LLMs. Moreover, we identify best practices for responsible LLM development, offering recommendations for transparency, collaboration, and awareness of research impacts. Beyond providing a view of the current progress of LLM research, these potential directions can further assist in the growth and positive impacts of LLMs.

\bibliography{sn-bibliography}

\end{document}